%% file: main_arxiv.tex
\documentclass[sigconf,screen,nonacm]{acmart}

\AtBeginDocument{%
  }

\usepackage{cleveref,stfloats,array,pifont,enumitem}
\usepackage[export]{adjustbox}
\usepackage[normalem]{ulem}
\Crefname{figure}{Fig.}{Figs.}
\Crefname{table}{Tab.}{Tabs.}
\Crefname{section}{Sec.}{Secs.}
\Crefname{equation}{Eq.}{Eqs.}
\Crefname{appendix}{Appx.}{Appx.}

\newcommand{\cmark}{\ding{51}}
\newcommand{\xmark}{\ding{55}}
\newcommand{\thirdbest}[1]{\uwave{#1}}

\newcommand{\METHODPLACEHOLDER}{PhyEdit}
\newcommand{\TITLEPLACEHOLDER}{\METHODPLACEHOLDER{}: Towards Real-World Object Manipulation via Physically-Grounded Image Editing}
\newcommand{\DATASETPLACEHOLDER}{RealManip-40K}
\newcommand{\BENCHMARKPLACEHOLDER}{ManipEval}

\begin{document}

\renewcommand\footnotetextcopyrightpermission[1]{}

\title{\TITLEPLACEHOLDER{}}

\author{Ruihang Xu}
\orcid{0009-0009-5106-0971}
\affiliation{%
  \institution{Zhejiang University}
  \city{Hangzhou}
  \country{China}
}
\email{ruihangxu@zju.edu.cn}

\author{Dewei Zhou}
\orcid{0000-0003-3125-046X}
\affiliation{%
  \institution{Zhejiang University}
  \city{Hangzhou}
  \country{China}
}
\email{zdw1999@zju.edu.cn}

\author{Xiaolong Shen}
\orcid{0000-0001-8838-611X}
\affiliation{%
  \institution{Zhejiang University}
  \city{Hangzhou}
  \country{China}
}
\email{sxlongcs@zju.edu.cn}

\author{Fan Ma}
\orcid{0000-0002-4131-1222}
\correspondingauthor
\affiliation{%
  \institution{Zhejiang University}
  \city{Hangzhou}
  \country{China}
}
\email{mafan@zju.edu.cn}

\author{Yi Yang}
\orcid{0000-0002-0512-880X}
\affiliation{%
  \institution{Zhejiang University}
  \city{Hangzhou}
  \country{China}
}
\email{yangyics@zju.edu.cn}

\renewcommand{\shortauthors}{Ruihang Xu, Dewei Zhou, Xiaolong Shen, Fan Ma, and Yi Yang}

\begin{abstract}
  \input{sections/00-abstract}
\end{abstract}



\begin{teaserfigure}
  \centering
  \includegraphics[width=0.9\textwidth]{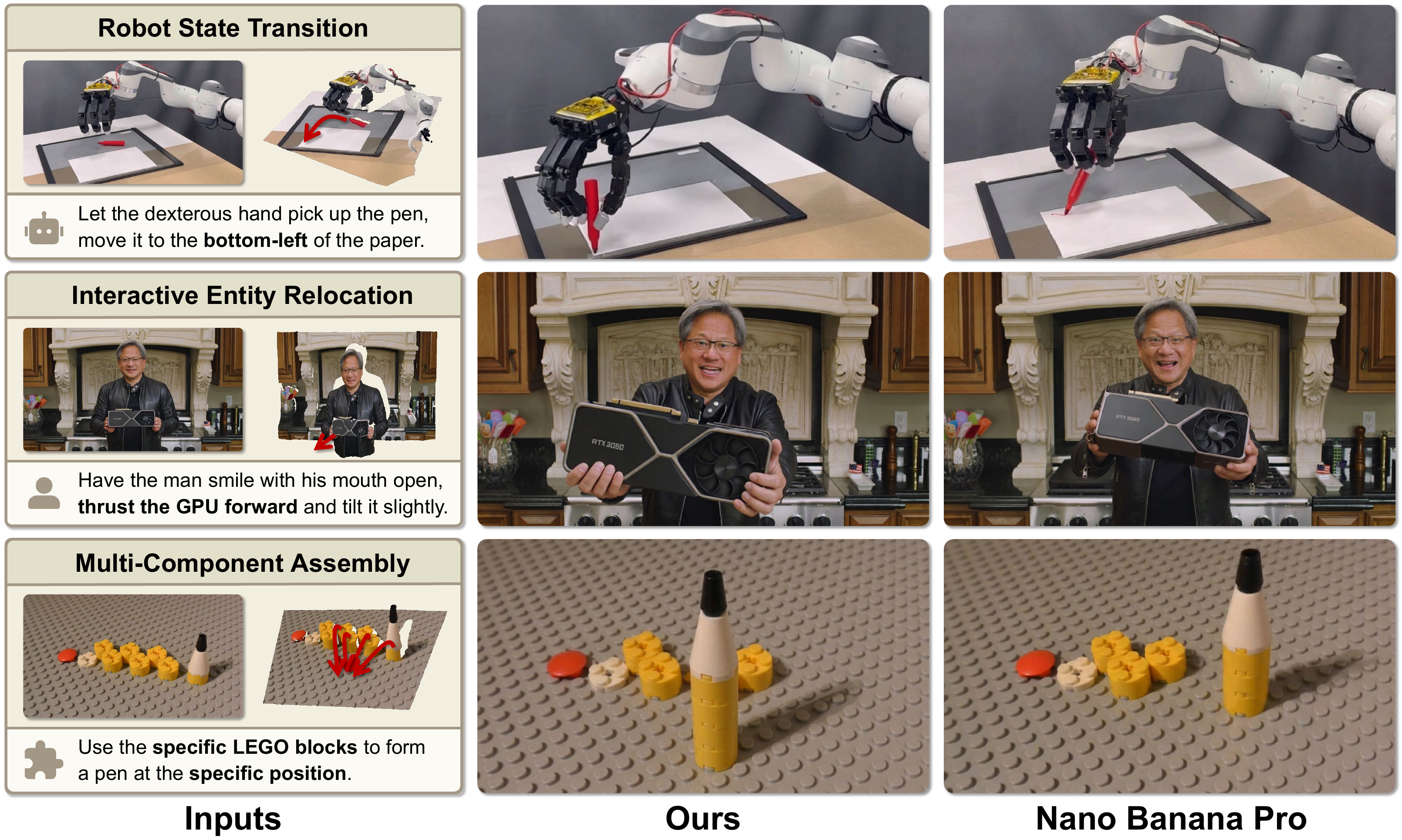}
  \caption{Comparison of image editing results on three physical scenarios between our PhyEdit and Nano Banana Pro~\cite{google-deepmind-2025-nanobanana-pro}}
  \Description{A three-row comparison with inputs and transformation instructions at left, PhyEdit outputs in the center, and Nano Banana Pro outputs at right. The rows show a robot hand moving a red pen to the lower-left of a sheet of paper, a smiling man thrusting and slightly tilting a graphics card, and selected LEGO pieces assembled into a pen at a specified position.}
  \label{fig:teaser}
\end{teaserfigure}

\maketitle

\setlist[itemize]{leftmargin=*, labelindent=\parindent}
\input{sections/01-introduction.tex}
\input{sections/02-relatedwork.tex}
\input{sections/03-method.tex}
\input{sections/04-experiments.tex}
\input{sections/05-conclusion.tex}

\begin{acks}
This work was supported by the Fundamental and Interdisciplinary Disciplines Breakthrough Plan of the Ministry of Education of China (JYB2025XDXM103).
\end{acks}

\bibliographystyle{ACM-Reference-Format}
\bibliography{main}

\clearpage
\appendix
\input{sections/06-appendix.tex}

\end{document}

%% file: sections/00-abstract.tex
Achieving physically accurate object manipulation in image editing is essential for its potential applications in interactive world models. However, existing visual generative models often fail at precise spatial manipulation, resulting in incorrect scaling and positioning of objects. This limitation primarily stems from the lack of explicit mechanisms to incorporate 3D geometry and perspective projection. To achieve accurate manipulation, we develop PhyEdit, an image editing framework that leverages explicit geometric simulation as contextual 3D-aware visual guidance. By combining this plug-and-play 3D prior with joint 2D--3D supervision, our method effectively improves physical accuracy and manipulation consistency. To support this method and evaluate performance, we present a real-world dataset, RealManip-40K, for 3D-aware object manipulation featuring paired images and depth annotations. We also propose ManipEval, a benchmark with multi-dimensional metrics to evaluate 3D spatial control and geometric consistency. Extensive experiments show that our approach outperforms existing methods, including strong closed-source models, in both 3D geometric accuracy and manipulation consistency. Code is available at \url{https://github.com/nenhang/PhyEdit}.

%% file: sections/01-introduction.tex
\section{Introduction}
Diffusion models have achieved impressive results in image generation and editing. Recent visual generative models~\cite{google-deepmind-2025-nanobanana-pro,qwen-team-2026-qwenimage2,openai-2025-chatgpt-images,bytedance-seed-2026-seedream50lite,flux-2-2025} perform seamless editing such as manipulating an object at the 2D level. These general models and those specialized for object manipulation~\cite{jiang2024actenhancedobjectmanipulation,yu2025objectmovergenerativeobjectmovement, zhang2025gooddrag, jiang2025pixelman,shi2024lightningdrag} execute precise spatial translation when the prompt contains detailed coordinate instructions (e.g., `move the object to position $(x, y)$'). However, 2D-level manipulation is often insufficient for emerging real-world applications. For instance, object-centric robotic manipulation guided by visual priors~\cite{pang2025imagegenerationvisualplanner,zhu2025lavamanlearningvisualaction,Black2023ZeroShotRM} requires precise physical manipulation in 3D space. This demands models to act as interactive world models capable of rendering 3D-aware and geometrically consistent state transitions. Achieving such precise rendering remains a significant challenge.

The limitations of existing models in acting as reliable state renderers stem from two main aspects. \textbf{(1) Lack of Explicit 3D and Physical Control.} Recent large visual generative models attempt to learn world dynamics from visual data~\cite{huang2025vid2world0,bruce2024genie,kang2025far}. However, without explicit mechanisms to incorporate perspective projection laws, these models often produce physically implausible behaviors, such as incorrect scale variations during object movement and inconsistent trajectories. In practical applications, this limitation prevents precise object manipulation, restricting users to vague textual prompts like ``further'' or ``closer'', or coarse scaling factors, making fine-grained geometric manipulation unattainable. \textbf{(2) Lack of High-Quality Real-World Datasets and Benchmarks.} Existing datasets for image  editing~\cite{zhang2025gooddrag,shi2024lightningdrag,cao2024instructmove} place little emphasis on 3D-aware spatial changes and real-world physical laws. Meanwhile, 3D asset datasets~\cite{yu2025objectmovergenerativeobjectmovement,michel2023object3ditlanguageguided3daware, objaverseXL} are primarily synthesized using 3D engines. They contain few non-rigid real-world objects. This limits the generalization of trained models in simulating the real world. Furthermore, related benchmarks~\cite{jiang2024actenhancedobjectmanipulation,zhang2025gooddrag,shi2023dragdiffusion,michel2023object3ditlanguageguided3daware} typically evaluate 2D metrics and overlook depth accuracy. To facilitate research in 3D image editing, both a high-quality real-world dataset and a benchmark with appropriate 3D evaluation metrics are required.

To bridge this gap, we introduce \textbf{\METHODPLACEHOLDER{}}, taking a step towards real-world object manipulation by framing 3D-aware editing as generative state rendering. Driven by explicit 3D movement instructions from either users or interactive systems, our DiT-based approach handles geometric displacement while naturally accounting for implicit physical effects such as deformations and occlusions. Specifically, our framework features a plug-and-play contextual 3D-aware visual guidance module to inject geometric priors, and a non-intrusive joint loss strategy utilizing both 2D and 3D supervision to improve spatial accuracy.

Along with the framework, we propose \textbf{\DATASETPLACEHOLDER{}}, a high-quality real-world dataset tailored for 3D-aware object manipulation. This dataset provides real-world image pairs demonstrating physical object manipulation in 3D space, along with detailed depth annotations. We also design a specialized benchmark, \textbf{\BENCHMARKPLACEHOLDER{}}, with representative metrics to evaluate spatial control accuracy.

In summary, our key contributions are threefold:

\begin{itemize}
    \item \textbf{Method}: We develop \textbf{\METHODPLACEHOLDER{}}, a DiT-based framework designed for physically-accurate image editing. It utilizes a contextual 3D visual guidance module and 2D-3D joint supervision to improve manipulation accuracy.
    \item \textbf{Dataset and Benchmark}: We introduce \textbf{\DATASETPLACEHOLDER{}} and \textbf{\BENCHMARKPLACEHOLDER{}} for training and evaluating real-world 3D-aware object manipulation.
    \item \textbf{SOTA Performance}: Our approach achieves superior performance on \BENCHMARKPLACEHOLDER{} compared to existing baselines in 3D manipulation precision. Ablation studies confirm the effectiveness of our design choices.
\end{itemize}

%% file: sections/02-relatedwork.tex
\section{Related Work}

\subsection{Object Manipulation}
In the context of interactive world models, manipulating the spatial arrangement of objects within an image can be viewed as rendering a state transition. 

\noindent\textbf{2D Object Manipulation.}
To achieve such transitions, existing approaches operating in the 2D plane generally fall into two categories: drag-based methods and explicit manipulation. Drag-based and point-conditioned methods~\cite{pan2023draggan,shi2023dragdiffusion,ling2024freedrag,zhang2025gooddrag,zhou2026metapoint} allow users to shift visual content through specified control points, but they typically require time-consuming per-image optimization during inference. Alternatively, explicit manipulation and region-specific refinement pipelines~\cite{jiang2024actenhancedobjectmanipulation,jiang2025pixelman,duan2025diffoom,zhou2026refineanything} process target subjects or selected regions through translation, refinement, and background inpainting, yet they often struggle with complex occlusions and multi-object interactions. Although these paradigms effectively edit objects within the image plane, they lack a structural understanding of 3D scene geometry.

\noindent\textbf{3D Object Manipulation.}
To address this limitation, another line of research tackles manipulation from a 3D perspective. Early methods recover explicit 3D geometry from images or videos~\cite{zhao20253dobjectmanipulationsingle,shen2023global,hu2026tracehighfidelity3dscene,chen2023gaussianeditor,sam3dteam2025sam3d3dfyimages} before subsequent manipulation or rendering. In editing pipelines, final image quality depends heavily on the accuracy of the reconstructed geometry and lighting, and non-rigid deformations remain difficult. More recent works avoid heavy reconstruction by incorporating 3D spatial information or visual generative priors directly into the editing process. For instance, some methods apply 3D transformations within the latent space~\cite{sajnani2024geodiffuser,pandey2024diffusionhandles}, encode spatial or spatiotemporal information into tokens~\cite{neuralassets_2024,tan2025sweettok}, or use video generation to guide spatial movements~\cite{Rotstein2024PathwaysOT,yu2025objectmovergenerativeobjectmovement,wu2025chronoedit}. Our work follows this generative line of research but emphasizes physically-grounded simulation to achieve precise 3D object manipulation.

\subsection{Datasets and Benchmarks for Object Manipulation}
\label{sec:related_dataset}
Current real-world image editing datasets~\cite{cao2024instructmove,motionedit,chang2025bytemorph} mostly focus on motion or shape changes. They lack samples that clearly reflect physical laws, such as perspective-induced size variations caused by depth movements. On the other hand, 3D simulation datasets~\cite{ahmadyan2020objectronlargescaledataset,michel2023object3ditlanguageguided3daware} offer limited visual diversity and typically focus on a single, isolated object. They cannot reflect the complex interactions among multiple objects during real-world state transitions. For evaluation, existing benchmarks~\cite{pan2023draggan,shi2023dragdiffusion,michel2023object3ditlanguageguided3daware,jiang2024actenhancedobjectmanipulation} mostly measure image feature distances and general quality, rather than testing control accuracy in actual physical and geometric dimensions.

%% file: sections/03-method.tex
\section{Method}

\begin{figure*}[t]
  \centering
  \includegraphics[width=0.9\linewidth]{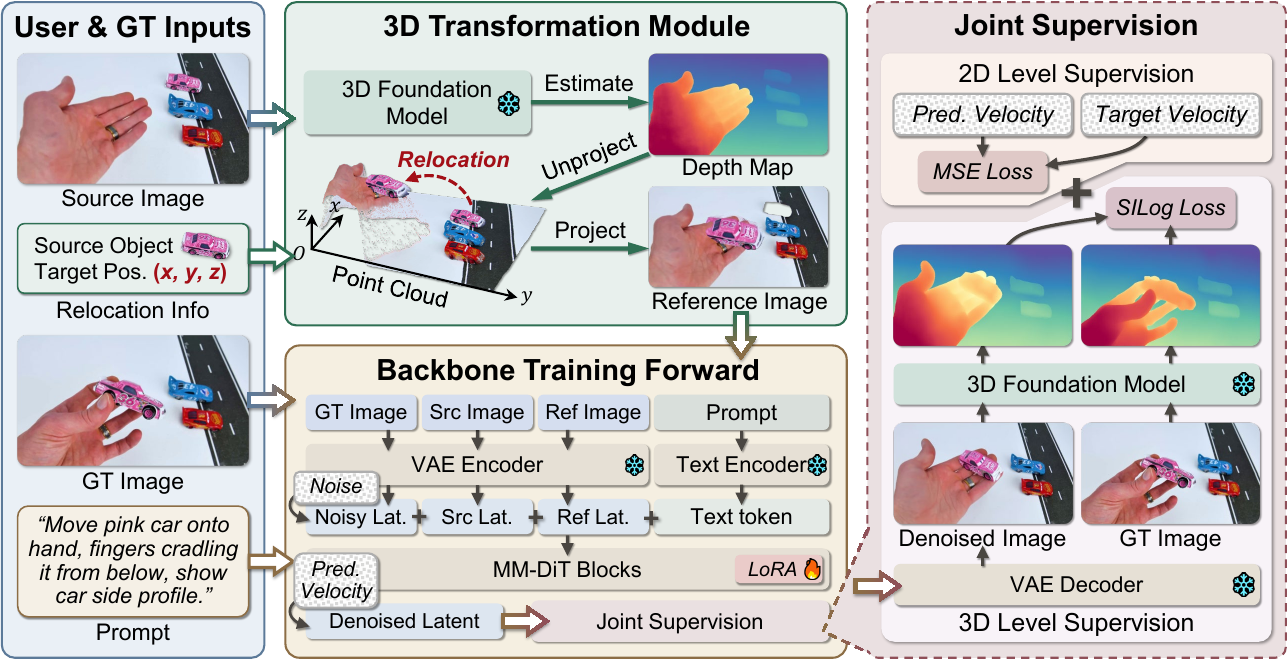}
  \caption{Overview of \METHODPLACEHOLDER{}. User and target inputs are first processed via the 3D transformation module. The resulting conditions are then fed into the backbone for the training forward pass, followed by 2D and 3D level joint supervision.}
  \label{fig:model_architecture}
  \Description{A left-to-right architecture diagram. User and target inputs provide a source image, a source object with a target 3D position, a target image, and a prompt. A frozen 3D foundation model estimates depth, unprojects a point cloud, relocates the object, and projects a reference image. The backbone encodes the target image, reference image, and prompt and processes them with LoRA-adapted MM-DiT blocks. Joint supervision combines flow-matching loss with pixel-level depth SILog loss computed after VAE decoding.}
\end{figure*}

\subsection{Preliminaries}
\noindent\textbf{Diffusion Transformer for Image Editing.} Diffusion Transformer~(DiT) models~\cite{Peebles2022DiT} are widely used in image generation and editing because they support flexible multi-modal conditioning. For one denoising step, the DiT input token sequence is
\begin{equation}
\mathbf{T}
=
\left[
\mathbf{T}_{\text{text}};
\mathbf{T}_{\text{latent}};
\mathbf{T}_{\text{cond}}
\right],
\label{eq:token_input}
\end{equation}
where $\mathbf{T}_{\text{text}}$ is the text-prompt embedding, $\mathbf{T}_{\text{latent}}$ is the noisy image latent token, and $\mathbf{T}_{\text{cond}}$ denotes additional condition tokens. Here $[\cdot;\cdot]$ means concatenation.

Given clean latent $\mathbf{z}_0$ and noise $\boldsymbol{\epsilon}\sim\mathcal{N}(0,\mathbf{I})$, flow matching defines a linear interpolation path parameterized by $t\in[0,1]$:
\begin{equation}
\mathbf{z}_t=(1-t)\mathbf{z}_0+t\boldsymbol{\epsilon}.
\end{equation}
The corresponding target velocity along this path is $\mathbf{v}^*=\boldsymbol{\epsilon}-\mathbf{z}_0$. The DiT is trained to predict this velocity:
\begin{equation}
\mathcal{L}_{\text{flow}}
=
\mathbb{E}_{t,\mathbf{z}_0,\boldsymbol{\epsilon}}
\left[
\left\|
v_{\theta}(\mathbf{T},t)-(\boldsymbol{\epsilon}-\mathbf{z}_0)
\right\|_2^2
\right].
\label{eq:diffusion_loss}
\end{equation}
Here $v_{\theta}$ is the predicted velocity, and the loss matches it to the target velocity $\mathbf{v}^*$.

\noindent\textbf{Multi-View 3D Foundation Model.} 
A multi-view 3D foundation model jointly predicts scene geometry and camera parameters from a set of input images. Recent methods such as VGGT~\cite{wang2025vggt}, Pi3~\cite{wang2025pi}, and Depth-Anything-3~\cite{depthanything3} adopt a Transformer followed by some prediction heads~\cite{Ranftl2021dpt}. Given $N$ input images $\{I_i\}_{i=1}^{N}$, each image is encoded and paired with a learnable camera token; the full sequence is processed jointly:
\begin{equation}
\left[\{z_{\text{cam},i}\};\{\mathbf{Z}_i\}\right]
=
F\!\left(\left[\{x_{\text{cam},i}\};\{E(I_i)\}\right]\right),\quad i=1,\dots,N.
\label{eq:threed_model_input}
\end{equation}
$E(\cdot)$ is the shared image encoder, $x_{\text{cam},i}$ is the learnable camera token for view $i$, and $F$ is the cross-view transformer. It outputs per-view camera features $z_{\text{cam},i}$ and visual tokens $\mathbf{Z}_i$. Prediction heads output depth and camera pose for each view:
\begin{equation}
D_i=H_d(\mathbf{Z}_i),\qquad (R_i,t_i)=H_{\text{cam}}(z_{\text{cam},i}),
\label{eq:threed_model_forward}
\end{equation}
where $H_d$ and $H_{\text{cam}}$ are shared prediction heads, $D_i$ is the depth map for view $i$, and $(R_i,t_i)$ is its camera rotation and translation.

\begin{figure*}[t]
  \centering
  \includegraphics[width=\linewidth]{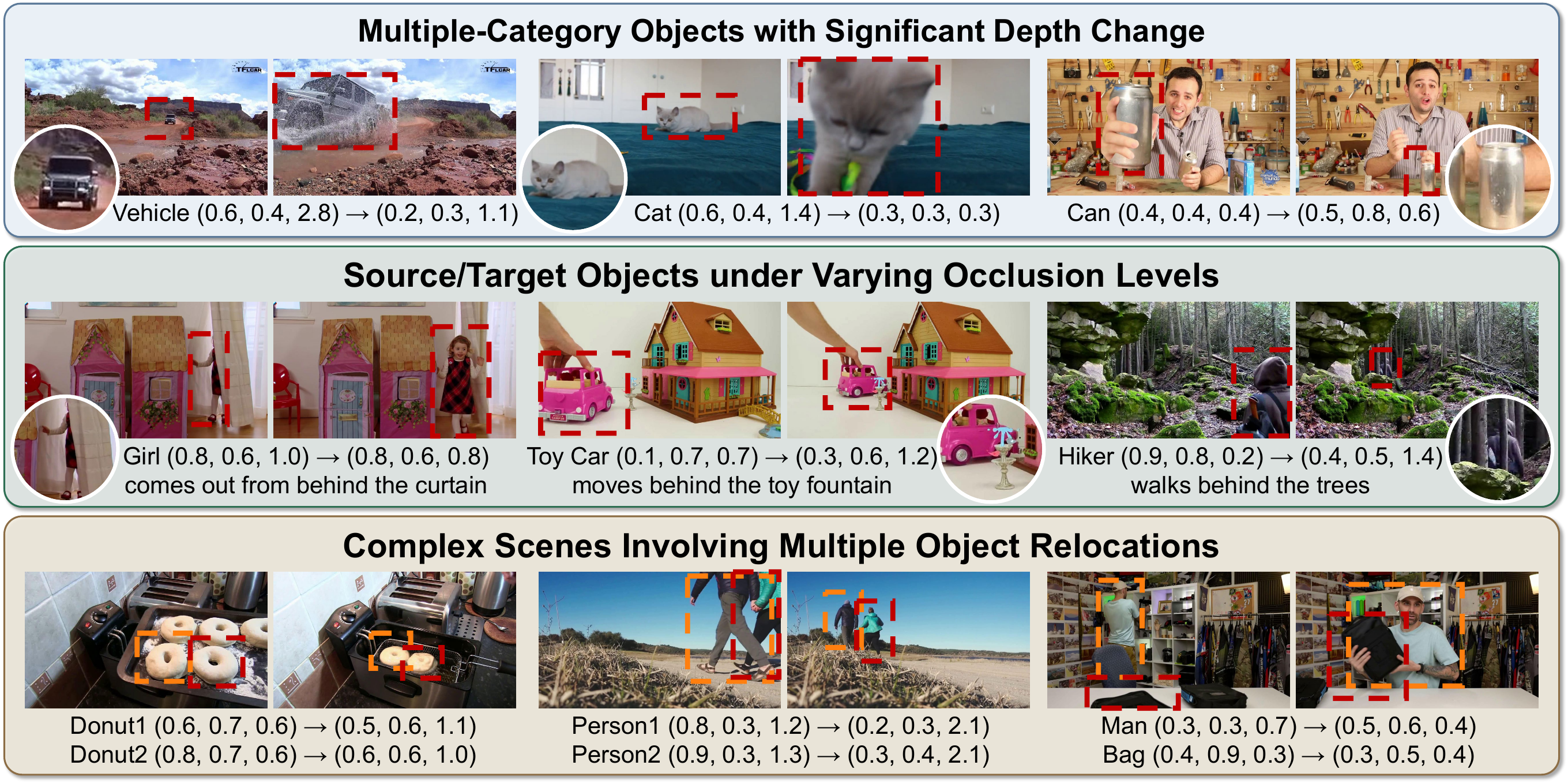}
  \caption{Samples from \DATASETPLACEHOLDER{}. The objects are annotated with {\color[HTML]{C00000}red} and {\color[HTML]{FF7F0F}orange} boxes. The normalized 3D object coordinates $(x, y, z)$ follow an image-aligned coordinate system with the origin at the top-left-near corner.}
  \label{fig:dataset_samples}
  \Description{Nine source-target examples arranged in three groups. The top group shows vehicle, cat, and can motions with large depth changes; the middle group shows a girl, toy car, and hiker becoming more or less occluded; and the bottom group shows paired relocations of donuts, people, and a man with a bag. Red and orange boxes identify manipulated objects, and each example lists source and target 3D coordinates.}
\end{figure*}

\subsection{Model Architecture}

\noindent\textbf{Overall Architecture.} 
To use 3D priors effectively for object manipulation, we combine a DiT editing backbone with a 3D foundation model (\Cref{fig:model_architecture}). The framework has three parts: (1) a 3D transformation module that generates a depth-aware preview, (2) the DiT denoising backbone, and (3) a joint training loss in 2D latent and 3D depth spaces.

\noindent\textbf{3D Transformation Module.}
\label{sec:3d_transformation_module}
Given source image $I_{\text{src}}$, object mask $M_o$ (manual or automatic), and transition vector $\Delta\mathbf{p}_o$, we first predict depth $D$ and camera pose ($R, t$), then edit the object directly in 3D space:
\begin{gather}
\mathbf{P}_{o}=\operatorname{Unproj}(I_{\text{src}}, M_{o},D,R,t),\\
\mathbf{P}_{o}'=\mathbf{P}_{o}+\Delta\mathbf{p}_o,\\
I_{\text{prev}}=\operatorname{Proj}(\mathbf{P}'_o,R,t).
\end{gather}
The preview image $I_{\text{prev}}$ is used as an additional condition for DiT; it does not replace the source image. The source remains the native visual input of the Qwen-Image-Edit backbone, while the preview supplies explicit geometric guidance. Inspired by previous work~\cite{xu2025contextgen,zhou2024migc,zhou2024migc++,zhou2025dreamrenderer}, this design naturally supports multi-object manipulation without iterative editing.

\noindent\textbf{Backbone Training Forward.}
During training, the DiT input combines text, noisy latent, source-image tokens, and preview tokens:
\begin{gather}
\mathbf{T}
=
\left[
\mathbf{T}_{\text{text}};
\mathbf{T}_{\text{latent}};
\mathbf{T}_{\text{src}};
\mathbf{T}_{\text{prev}}
\right],\\
v_{\theta}(\mathbf{T},t)
=
\operatorname{MM\mbox{-}DiT}(\mathbf{T},t).
\end{gather}
Here $\mathbf{T}_{\text{src}}$ and $\mathbf{T}_{\text{prev}}$ denote the source-image and transformed-preview tokens, respectively.

\noindent\textbf{Joint Supervision.}
\label{joint_loss_system}
Latent-space flow-matching loss alone is often insufficient for 3D manipulation, because it emphasizes appearance reconstruction more than geometric correctness. We therefore add a depth-space supervision term.

We first estimate the clean latent and decode the edited image:
\begin{gather}
\hat{\mathbf{z}}_0
=\mathbf{z}_t-t\,v_{\theta}(\mathbf{T},t)
,\\
\hat{I}_{\text{edit}}
=
\operatorname{Dec}(\hat{\mathbf{z}}_0),\\
\mathcal{L}_{\text{depth}}
=
\operatorname{SILog}\!\left(D_{\text{edit}},D_{\text{tgt}}\right),
\end{gather}
where $D_{\text{edit}}$ and $D_{\text{tgt}}$ are depth maps of the edited and target images. We then use the scale-invariant logarithmic loss~\cite{Eigen2014DepthMP}:
\begin{equation}
\begin{split}
\operatorname{SILog}(D_{\text{pred}},D_{\text{tgt}})
&=
\frac{1}{N}\sum_{i=1}^{N}\Delta_i^2
-\frac{1}{N^2}\!\left(\sum_{i=1}^{N}\Delta_i\right)^{\!2},\\
\Delta_i&=\log D_{\text{pred}}(i)-\log D_{\text{tgt}}(i),
\end{split}
\end{equation}

where $N$ is the number of valid depth pixels. The final objective is
\begin{equation}
\mathcal{L}
=
\mathcal{L}_{\text{flow}}
+\lambda_d\mathcal{L}_{\text{depth}},
\end{equation}
where $\lambda_d$ balances latent and depth supervision.

Compared with methods that require heavy architecture changes to inject 3D priors~\cite{neuralassets_2024,sajnani2024geodiffuser,zhang20243DitScene}, our approach only adds lightweight conditioning and training design. It can be plugged into different DiT-based editors with minimal modification.

\subsection{Dataset Construction}

\noindent\textbf{Dataset Overview.}
As discussed in \Cref{sec:related_dataset}, existing datasets are not sufficient for 3D-aware object manipulation. We therefore build \textsc{\DATASETPLACEHOLDER{}}, a real-world dataset of paired images $(I_{\text{src}},I_{\text{tgt}})$ where one or more objects are manipulated in 3D space, especially along the depth axis. Each pair provides depth maps, object masks, and representative 3D object coordinates.

\begin{figure*}[t]
  \centering
  \includegraphics[width=\linewidth]{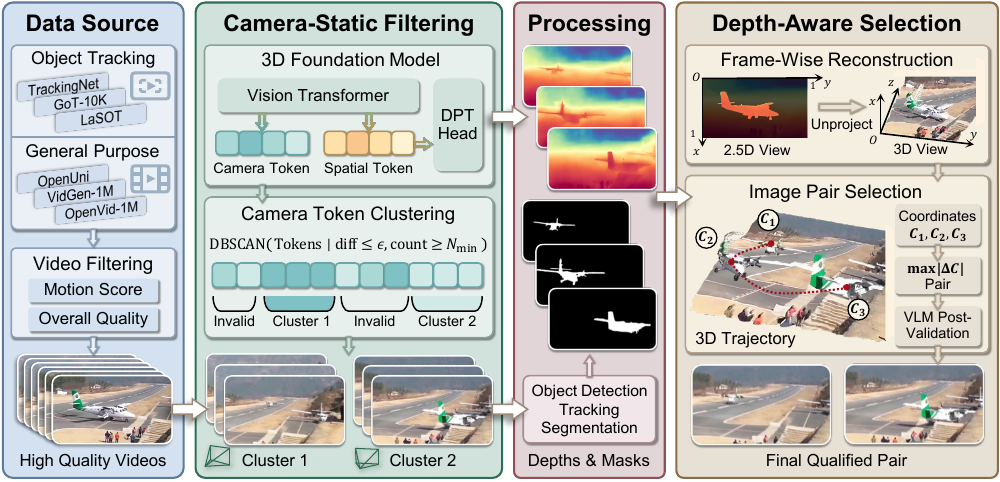}
  \caption{Dataset Construction Pipeline. The workflow consists of four key stages: data source filtering, camera-static clip extraction, depth and mask processing, and depth-aware frame pair selection. At the bottom of the second column, the symbols \includegraphics[height=1.2em, valign=c]{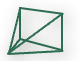} and \includegraphics[height=1.2em, valign=c]{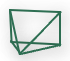} represent 3D viewing frustums, denoting that Cluster 1 and Cluster 2 have different camera viewpoints.}  
  \label{fig:dataset_construction}
  \Description{A four-stage pipeline from left to right. Videos from tracking and general-purpose datasets are filtered by motion and quality. A 3D foundation model clusters camera tokens to retain camera-static clips. Object detection, tracking, segmentation, and depth estimation produce masks and depth maps. Frame-wise 3D reconstruction then selects the image pair with the largest valid object-coordinate change, followed by VLM validation.}
\end{figure*}

As shown in \Cref{fig:dataset_samples}, the dataset covers diverse scenes and object types. Several features make it suitable for 3D manipulation learning: 
(1) Each pair has an object with significant depth change. 
(2) The dataset includes samples with part of the object occluded, which emphasizes the near-far relationship within the scene. 
(3) The dataset contains pairs with multiple objects manipulated simultaneously, which encourages the model to learn complex spatial interactions. The construction pipeline is shown in \Cref{fig:dataset_construction}. We detail the pipeline in the following parts.

\noindent\textbf{Data Source.}
\label{data_source_selection}
Data quality is critical: low-quality sources can weaken the supervision signal and hurt generation quality. We collect videos from high-resolution and minimally overlapping datasets, including \textit{OpenVid-1M}~\cite{nan2024openvid}, \textit{VIDGEN-1M}~\cite{tan2024vdgen-1m}, and \textit{OpenUni-Dataset}~\cite{huang2026unityvideo}. We also include object-tracking datasets (\textit{LaSOT}~\cite{fan2020lasothighqualitylargescalesingle}, \textit{GoT-10K}~\cite{Huang2021got10k} and \textit{TrackingNet}~\cite{muller2018trackingnetlargescaledatasetbenchmark}) as supplements. All clips are filtered by video quality assessment (VQA)~\cite{wu2022fastquality} and motion estimation~\cite{wang2024mattersnovelvisualphysicsbased}.

\noindent\textbf{Camera-Static Filtering.}
Reliable manipulation supervision requires near-static cameras in each pair. Instead of using optical flow~\cite{10.1007/3-540-45103-X_50} or hand-crafted feature matching~\cite{lowe2004distinctive}, we use \textit{camera token clustering}. The camera token $z_{\text{cam}}$ in \Cref{eq:threed_model_forward} is predicted by \textit{3D foundation models}. We cluster these tokens with DBSCAN~\cite{ester1996density} and keep high-density clusters as camera-static clips:
\begin{equation}
\mathcal{C}
=
\mathrm{DBSCAN}\!\left(
\{z_{\text{cam}}^{(k)}\}_{k=1}^{K};
\epsilon,\,
N_{\min}
\right),
\end{equation}
where $\epsilon$ is the neighborhood radius and $N_{\min}$ is the minimum number of samples per cluster.

\noindent\textbf{Depths \& Masks Processing.}
For each selected cluster, we predict per-frame depth with the 3D foundation model, detect main objects on key frames with an open-world detector~\cite{yolo26_ultralytics}, and propagate masks through the clip using a video object tracker~\cite{carion2025sam3segmentconcepts}.

\noindent\textbf{Depth-Aware Selection.}
To select pairs with clear 3D manipulation, we unproject each masked frame ($I^{(k)} * M^{(k)}$) into 3D using depth $D^{(k)}$ and camera pose ($R^{(k)}, t^{(k)}$), then estimate an object representative coordinate with a coordinate-wise median:
\begin{gather}
\mathbf{c}_{o}^{(k)}
=
\operatorname{Median}\!\left(
\operatorname{Unproj}(I^{(k)}*M_o^{(k)},D^{(k)},R^{(k)},t^{(k)})
\right),\\
d_{o}^{(i,j)}
=
\left\|
\mathbf{c}_{o}^{(i)}-\mathbf{c}_{o}^{(j)}
\right\|_2.
\end{gather}
We choose the frame pair ($i, j$) with the largest distance $d_o^{(i,j)}$. For short clips, we use the first and last frames to reduce overhead. Finally, we apply depth-threshold filtering and VLM verification~\cite{qwen3.5} to remove samples with incorrect masks or insufficient 3D displacement. Each pair is further annotated with expanded prompts that describe non-geometric appearance changes (e.g., color or background variation), facilitating better text-image alignment for subsequent training.

%% file: sections/04-experiments.tex
\section{Experiments}
\label{sec:experiments}
\providecommand{\thirdbest}[1]{\textit{#1}}

\begin{table*}[t]
    \caption{Quantitative comparison on \BENCHMARKPLACEHOLDER{}. All metrics are linearly normalized to $[0,100]$. Methods are sorted by Chamfer distance in descending order. Proprietary commercial models are marked with $\dagger$. }
    \centering
    \begin{tabular}{c|*{9}{>{\centering\arraybackslash}w{c}{1.2cm}}}
    \toprule
     \textbf{Method} & DIoU$\uparrow$ & Mask IoU$\uparrow$ & AbsRel$\downarrow$ & $\delta_{1.25}$$\uparrow$ & Chamfer$\downarrow$ & Centroid$\downarrow$ & RA-DINO$\uparrow$ & DeQA$\uparrow$ & Phys-VLM$\uparrow$ \\
    \midrule
    GeoDiffuser & 51.72 & \underline{20.10} & 67.64 & 31.40 & 73.41 & 64.56 & 14.78 & 56.72 & 54.62 \\
    DiffusionHandles & 56.22 & 18.88 & 57.36 & 31.73 & 73.00 & 59.65 & 17.08 & 61.83 & 46.90 \\
    Move-and-Act & 46.89 & 10.97 & 68.47 & 29.30 & 64.59 & 63.62 & 13.01 & 60.79 & 60.85 \\
    OBJect 3DIT & 46.63 & 10.18 & 66.93 & 32.75 & 62.30 & 64.14 & 14.31 & 48.01 & 66.65 \\
    GoodDrag & 51.07 & 13.29 & 69.60 & 34.19 & 51.27 & 58.39 & 20.85 & 74.08 & 85.08 \\
    PixelMan & 49.48 & 11.49 & 66.46 & 35.10 & 50.18 & 57.13 & 21.17 & 72.79 & 73.60 \\
    Qwen-Image-Edit & 53.28 & 13.80 & 64.63 & 40.61 & 46.31 & 52.48 & 26.09 & \thirdbest{75.67} & 90.55 \\
    LightningDrag & 53.81 & 18.90 & 57.07 & 35.99 & 45.76 & 53.57 & 21.92 & 73.60 & 88.60 \\
    ChronoEdit & 48.92 & 9.26 & 65.80 & 36.35 & 45.02 & 54.40 & 21.73 & 75.29 & \underline{92.15} \\
    GPT-Image-1.5$^\dagger$ & 52.33 & 11.34 & 70.95 & 36.50 & 39.79 & 52.78 & 23.80 & \underline{75.82} & 88.68 \\
    Qwen-Image-2.0-Pro$^\dagger$ & \thirdbest{56.48} & 15.60 & \thirdbest{55.93} & \thirdbest{41.39} & \thirdbest{29.64} & \thirdbest{43.48} & \thirdbest{31.04} & 68.26 & 82.91 \\
    Nano Banana Pro$^\dagger$ & \underline{59.97} & \thirdbest{18.93} & \underline{55.02} & \underline{46.11} & \underline{25.33} & \underline{35.62} & \underline{34.77} & \textbf{77.48} & \thirdbest{91.06} \\
    \midrule
    Ours & \textbf{65.33} & \textbf{27.20} & \textbf{49.53} & \textbf{51.08} & \textbf{18.93} & \textbf{32.12} & \textbf{36.91} & 75.48 & \textbf{93.72} \\
    \bottomrule
    \end{tabular}
    
    \label{tab:baseline_comparison}
\end{table*}

\subsection{Training Details}
We use Qwen-Image-Edit~\cite{qwen-image-edit-2511} as the DiT editing backbone and Depth-Anything-3~\cite{depthanything3} as the 3D foundation model. We fine-tune the DiT with LoRA~\cite{hu2022lora} (rank $r=128$), mainly on self-attention layers. Training is conducted on 4$\times$ NVIDIA RTX PRO 6000 GPUs, with total batch size 64 for 10K steps. We use AdamW~\cite{Loshchilov2017DecoupledWD} with learning rate $1\times 10^{-4}$ and weight decay $0.01$. The depth-loss weight is set to $\lambda_d=0.1$.

\subsection{Benchmark and Metrics}

\noindent\textbf{Data.} Since there is no widely accepted benchmark for this task, we build an evaluation set with 200 image pairs and about 320 individual objects. It covers diverse scenes, object categories, and object scales. Half of the pairs contain a single manipulated object, and the other half contain multiple manipulated objects. Each pair includes depth annotations and object-level labels.

\noindent\textbf{Metrics.} We report metrics from five aspects. 

\paragraph{1. 2D Spatial Accuracy. }
\begin{itemize}
    \item \textbf{DIoU}~\cite{zheng2020diou}: Distance IoU between predicted and ground-truth boxes.
    \item \textbf{Mask IoU}: IoU between predicted and ground-truth masks.
\end{itemize}

\paragraph{2. Depth Accuracy}
Given predicted depth $D^{\text{pred}}$, target depth $D^{\text{tgt}}$, and valid object pixels $\Omega$:
\begin{itemize}
    \item \textbf{AbsRel}~\cite{Eigen2014DepthMP}:
    \begin{equation}
    \operatorname{AbsRel}
    =
    \frac{1}{|\Omega|}
    \sum_{i\in\Omega}
    \frac{|D^{\text{pred}}_i-D^{\text{tgt}}_i|}{D^{\text{tgt}}_i}.
    \end{equation}
    \item $\boldsymbol{\delta_{1.25}}$~\cite{Eigen2014DepthMP}: ratio of pixels satisfying
    $\max\!\left(\frac{D^{\text{pred}}_i}{D^{\text{tgt}}_i},\frac{D^{\text{tgt}}_i}{D^{\text{pred}}_i}\right)<1.25$.
\end{itemize}

\paragraph{3. 3D Manipulation Accuracy}
We reconstruct object point clouds and compare them with the target:
\begin{itemize}
    \item \textbf{Chamfer Distance}~\cite{Barrow1977ParametricCA}:
    \begin{equation}
    \begin{split}
    \operatorname{CD}(P^{\text{pred}},P^{\text{tgt}})
    &=
    \frac{1}{|P^{\text{pred}}|}
    \sum_{p\in P^{\text{pred}}}
    \min_{q\in P^{\text{tgt}}}\|p-q\|_2\\
    &\quad+
    \frac{1}{|P^{\text{tgt}}|}
    \sum_{q\in P^{\text{tgt}}}
    \min_{p\in P^{\text{pred}}}\|q-p\|_2.
    \end{split}
    \end{equation}
    We normalize point clouds by the valid-scene diagonal to make scores comparable.
    \item \textbf{Centroid Distance}:
    \begin{equation}
    \mathbf{c}^{\text{pred}}
    =
    \frac{1}{|P^{\text{pred}}|}
    \sum_{p\in P^{\text{pred}}}p,\qquad
    \mathbf{c}^{\text{tgt}}
    =
    \frac{1}{|P^{\text{tgt}}|}
    \sum_{q\in P^{\text{tgt}}}q,
    \label{eq:centroid}
    \end{equation}
    \begin{equation}
    \operatorname{CentroidDist}(P^{\text{pred}},P^{\text{tgt}})
    =
    \|\mathbf{c}^{\text{pred}}-\mathbf{c}^{\text{tgt}}\|_2.
    \end{equation}
\end{itemize}

\paragraph{4. Image Quality and Consistency}
\begin{itemize}
    \item \textbf{Relocation-Aware DINO Similarity}: DINO similarity~\cite{simeoni2025dinov3} penalized by relocation-vector errors.
    Let $\mathbf{v}^{\text{pred}}$ be decomposed into components parallel and orthogonal to $\mathbf{v}^{\text{tgt}}$:
    \begin{gather}
    e_{\parallel}
    =
    \frac{\|\mathbf{v}_{\parallel}-\mathbf{v}^{\text{tgt}}\|_2}
    {\|\mathbf{v}^{\text{tgt}}\|_2+\varepsilon},\qquad
    e_{\perp}
    =
    \frac{\|\mathbf{v}_{\perp}\|_2}
    {\|\mathbf{v}^{\text{tgt}}\|_2+\varepsilon},\\
    S_{\text{RA-DINO}}
    =
    S_{\text{DINO}}
    \cdot
    \exp\!\left(-\alpha e_{\parallel}-\beta e_{\perp}\right).
    \end{gather}
    We set $\alpha=1$ and $\beta=0.8$.
    \item \textbf{DeQA Score}~\cite{deqa_score}: general perceptual quality metric.
\end{itemize}

\paragraph{5. VLM Physical Plausibility}
\begin{itemize}
    \item \textbf{Phys-VLM}: VLM-based assessment of physical realism and global scene consistency (e.g., lighting/shadows, depth ordering, contacts/occlusions).
\end{itemize}

\subsection{Baselines}
We compare with five groups of baselines:\\
(1) \textbf{Drag-based}:
\textit{GoodDrag}~\cite{zhang2025gooddrag},
\textit{LightningDrag}~\cite{shi2024lightningdrag};\\
(2) \textbf{Explicit manipulation}:
\textit{Move-and-Act}~\cite{jiang2024actenhancedobjectmanipulation},
\textit{PixelMan}~\cite{jiang2025pixelman};\\
(3) \textbf{3D-aware editing}:
\textit{OBJect 3DIT}~\cite{michel2023object3ditlanguageguided3daware},
\textit{GeoDiffuser}~\cite{sajnani2024geodiffuser},
\textit{DiffusionHandles}~\cite{pandey2024diffusionhandles};\\
(4) \textbf{Video-prior}:
\textit{ChronoEdit}~\cite{wu2025chronoedit};\\
(5) \textbf{Commercial models}:
\textit{Nano Banana Pro}~\cite{google-deepmind-2025-nanobanana-pro},
\textit{GPT-Image-1.5}~\cite{openai-2025-chatgpt-images},
\textit{Qwen-Image-2.0-Pro}~\cite{qwen-team-2026-qwenimage2}.
For commercial systems, we use the strongest publicly available reasoning version.

Besides the methods supporting explicit 3D manipulation, we adjust the input for the other methods to perform a fairer comparison. Setting details are shown in the appendix.

\subsection{Comparison}

\begin{figure*}[t]
    \centering
    \includegraphics[width=\textwidth]{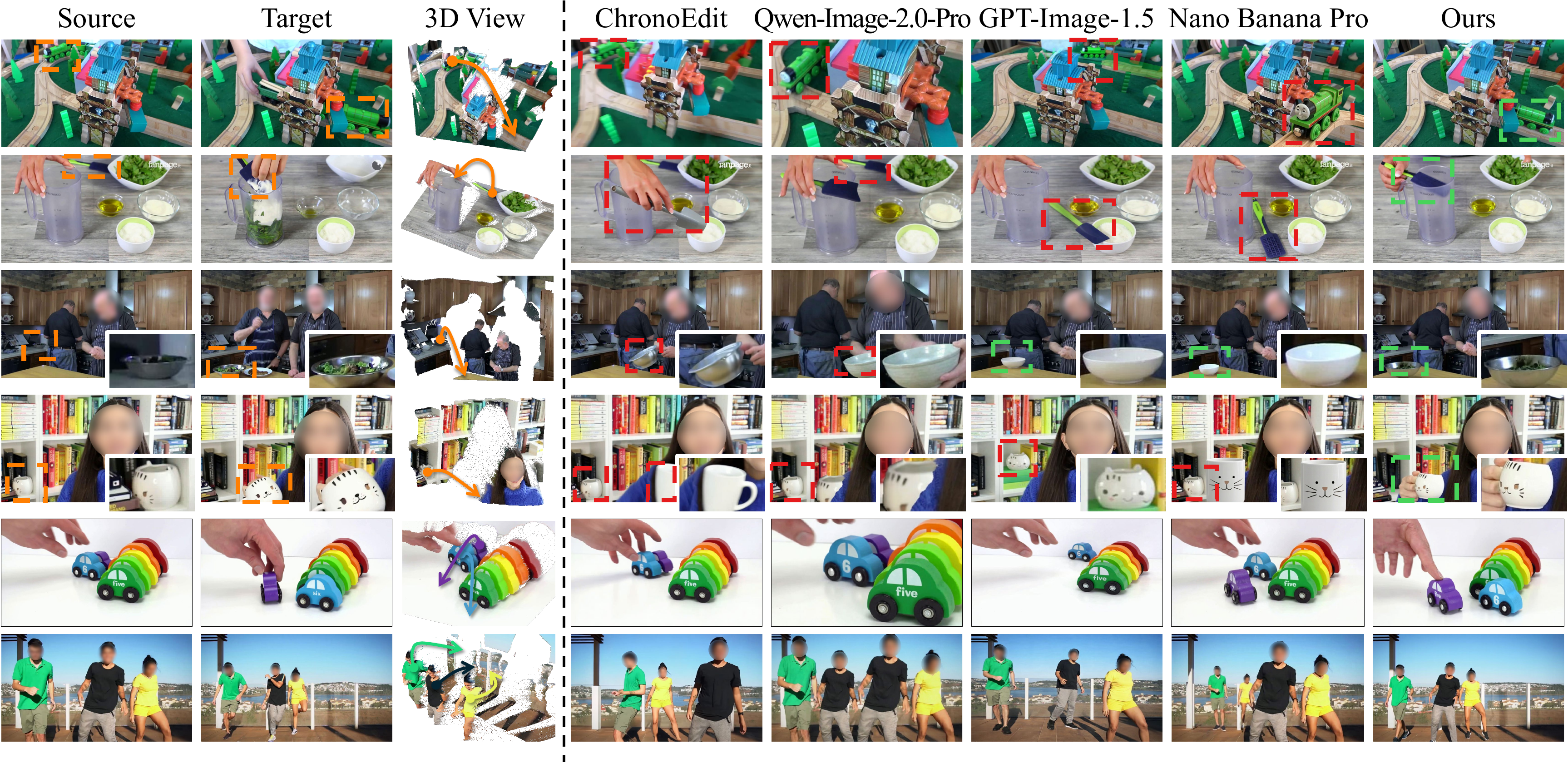}
    \caption{Qualitative comparison on \BENCHMARKPLACEHOLDER{}. The top two rows highlight manipulation accuracy and geometric consistency; the middle two rows focus on preserving object identity and fine details; the last two rows show multi-object manipulation. In the first four rows, source and target positions are annotated with {\color[HTML]{FF7F02}orange} boxes. Objects that are misplaced or still left at the source are marked with {\color[HTML]{E51B1D}red} boxes, and correctly manipulated objects are indicated by {\color[HTML]{47D45A}green} boxes.}
    \Description{A six-row qualitative comparison with columns for the source, target, 3D view, four baseline methods, and PhyEdit. The rows include moving a toy train around a building, inserting a spatula into a pitcher, relocating a metal bowl and a patterned mug, repositioning multiple toy cars, and changing the positions of several people. Orange boxes mark source and target regions, red boxes mark incorrect or incomplete baseline edits, and green boxes mark successful placements.}
    \label{fig:bench_demo}
\end{figure*}

\begin{figure*}[t]
    \centering
    \includegraphics[width=\textwidth]{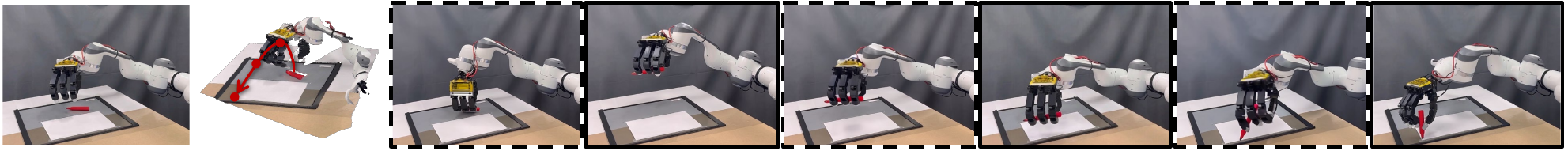}
    \caption{Continuous object manipulation along a trajectory. Given the initial image and a trajectory, \METHODPLACEHOLDER{} performs continuous editing (solid borders). Intermediate frames are further interpolated by a video model (dashed borders). Notably, this robotic arm is out of our training distribution, demonstrating our strong generalization.}
    \Description{A source image and a 3D trajectory are followed by six frames of a robot arm moving a red pen along a curved path toward the lower-left of a sheet of paper. Solid borders denote PhyEdit keyframes, and dashed borders denote interpolated frames.}
    \label{fig:continuous_object_manipulation}
\end{figure*}

\textbf{Quantitative results} on \BENCHMARKPLACEHOLDER{} are shown in \Cref{tab:baseline_comparison}. A consistent pattern is that existing methods with explicit 2D/3D control still lag behind strong commercial models on final manipulation accuracy, with notably larger gaps on depth and 3D geometry than on 2D overlap. Our method substantially narrows this gap and achieves the \textbf{best overall performance} across manipulation-related metrics, while also obtaining the highest RA-DINO score and the strongest Phys-VLM among all baselines, including closed-source commercial systems---indicating better physical plausibility and scene-level consistency after the move, beyond what raw layout or depth scores alone capture.

\textbf{Comparison with Commercial Baselines.} Against the strong commercial baseline, Nano Banana Pro, our method improves DIoU by \textbf{5.36} points (65.33 vs.\ 59.97), reduces Chamfer distance by \textbf{6.40} points (18.93 vs.\ 25.33), and improves RA-DINO by \textbf{2.14} points (36.91 vs.\ 34.77). Similar trends are also observed against other commercial systems, especially on geometry-sensitive metrics.

\textbf{Multi-Object Manipulation.} As detailed in the appendix, our lead over Nano Banana Pro grows from 5.19 to 6.58 in Chamfer distance and from 1.91 to 8.02 in $\delta_{1.25}$ from single- to multi-object scenarios. This demonstrates our superior capability to maintain precise 3D geometry in multi-object scenes.

\textbf{Image Quality Discussion.} The DeQA score is close to our base model Qwen-Image-Edit (75.48 vs.\ 75.67), suggesting that the gains in controllable manipulation and geometric consistency are achieved without noticeable loss of image quality. 

\noindent\textbf{Qualitative results} are shown in \Cref{fig:bench_demo}. The first two rows focus on manipulation accuracy and consistency. The middle two rows focus on preserving object identity and fine details. The last two rows focus on multi-object editing.

\textbf{Manipulation accuracy and consistency.} In the first two rows, our method places the toy train and the spatula at the correct 2D and depth locations. Most baselines either fail to move the object, leave a duplicate at the source location, or place the object at an incorrect depth. Our method also handles occlusions more effectively: the train remains partially occluded by the toy building, and the spatula is correctly inserted into the pitcher.

\textbf{Object identity and fine details.} In the third row, the target metal bowl is small. Several baselines did not manipulate the correct object or change its color or texture. In the fourth row, the target mug is also small and partially occluded, and baseline methods struggle to preserve the kitty pattern. 

\textbf{Multi-object editing.} In the last two rows, baseline methods often fail to edit all requested objects. Even the strongest commercial baseline, \textit{Nano Banana Pro}, handles only part of the targets, while our method edits all objects consistently.

\noindent\textbf{Continuous Object Manipulation.} Beyond single-step editing, \METHODPLACEHOLDER{} is capable of maintaining geometric consistency under continuous state transitions. As illustrated in \Cref{fig:continuous_object_manipulation}, given the initial state and spatial trajectory (and optional instructions), our model sequentially renders physically accurate keyframes at specified target points. These sparse frames can then be processed by interpolation models~\cite{wan2025wan} to synthesize a continuous, physically consistent object manipulation video.

\subsection{Ablation Studies}

\begin{figure}[t]
    \centering
    \includegraphics[width=0.45\textwidth]{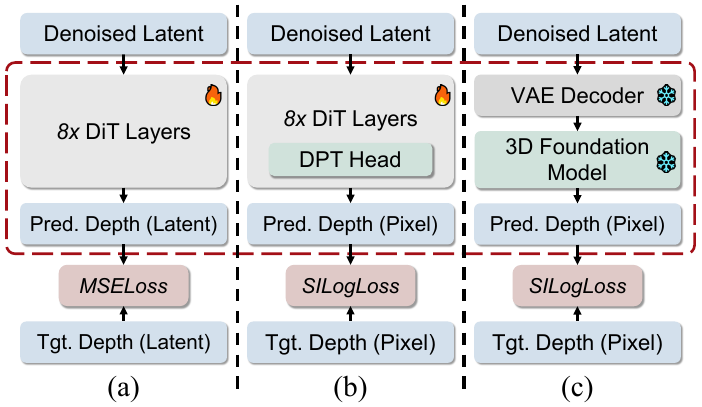}
    \caption{Architectures of different depth supervision methods. The core difference is highlighted by a {\color[HTML]{A30F17}red dashed} box.}
    \Description{Three parallel depth-supervision designs. The latent-to-latent design predicts a depth latent with eight trainable DiT layers and uses MSE loss. The latent-to-depth design adds a DPT head to predict pixel depth and uses SILog loss. The pixel-level design decodes the denoised latent with a frozen VAE, estimates depth with a frozen 3D foundation model, and applies SILog loss against target-image depth.}
    \label{fig:depth_head_ablation}
\end{figure}

\noindent\textbf{Depth Supervision Methods.} We compare three depth supervision variants, illustrated in \Cref{fig:depth_head_ablation}. 

\textbf{(a) Latent-to-Latent Supervision}: 
We train a latent-to-latent module with 8 DiT layers (similar parameter count to DA3~\cite{depthanything3}) to predict the latent of the depth map from the image latent.
\textbf{(b) Latent-to-Depth Supervision}:
We add a DPT head~\cite{Ranftl2021dpt} on top of the latent-to-latent module in \textbf{(a)}. It takes hidden features from several DiT layers and predicts the depth map without decoding the edited image to pixels.
\textbf{(c) Pixel-level Supervision}:
We decode the denoised latent and supervise the depth map of the decoded image (pixel-level alignment with the 3D prior).

\begin{table}[t]
    \centering
    \caption{Quantitative comparison of different depth supervision methods.}
    \label{tab:ablation_depth_supervision}
    \begin{tabular}{c*{4}{>{\centering\arraybackslash}w{c}{1.1cm}}}
        \toprule
         Method & DIoU$\uparrow$ & AbsRel$\downarrow$ & Chamfer$\downarrow$  & RA-DINO$\uparrow$ \\
        \midrule
        w/o Supervision & 62.37 & 50.92 & 24.52 & 33.53 \\
        Latent-to-Latent & 60.48 & 49.78 & 28.61 & 32.29 \\
        Latent-to-Depth & \underline{64.19} & \underline{49.73} & \underline{20.87} & \underline{35.93} \\
        Ours (pixel-level) & \textbf{65.33} & \textbf{49.53} & \textbf{18.93} & \textbf{36.91} \\
        \bottomrule
    \end{tabular}
\end{table}

\textbf{Quantitative results} are shown in \Cref{tab:ablation_depth_supervision}. All depth-supervised variants improve AbsRel over the model without supervision. The gains are around 1 point. However, differences become larger on 2D and 3D metrics. We observe: \textbf{1. The Latent-to-Latent variant} is weakest. It drops DIoU by 1.89 points relative to the no-supervision model, and also gives worse Chamfer and RA-DINO. This indicates that pure latent-level depth signals are not precise enough for accurate manipulation.
\textbf{2. The Latent-to-Depth} variant is stronger and improves all metrics over no supervision, but it is still below our \textbf{pixel-level} method (e.g., DIoU 64.19 vs.\ 65.33, Chamfer 20.87 vs.\ 18.93). Overall, the pixel-level strategy gives the best balance across 2D, depth, and 3D metrics.

\begin{figure}[t]
    \centering
    \includegraphics[width=0.4\textwidth]{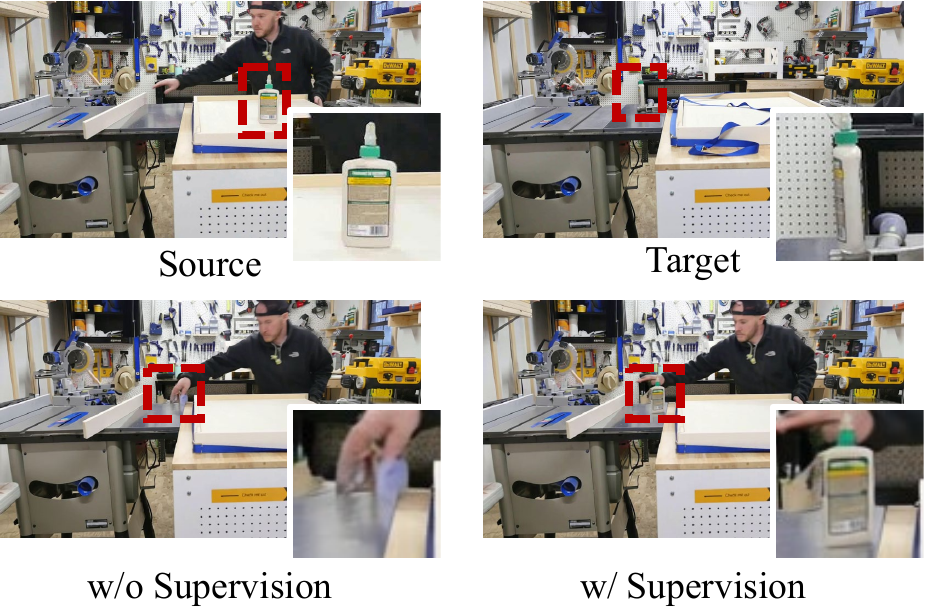}
    \caption{Qualitative comparison of methods with and without depth supervision.}
    \Description{A two-by-two example showing the source, target, output without depth supervision, and output with depth supervision. The task moves a small glue bottle from the right side of a workshop table to a target position near the saw. Without supervision the bottle is a blurred, distorted trace; with supervision it is recognizable and positioned close to the target. Enlarged crops highlight the small object.}
    \label{fig:supervision_ablation_demo}
\end{figure}

\textbf{Qualitative results} are shown in \Cref{fig:supervision_ablation_demo}. In the second row, we compare no depth supervision and pixel-level depth supervision under the same random seed. The target object is very small (smaller than one patch). Without depth supervision, the model only leaves a blurry trace near the target location. With pixel-level depth supervision, the object is generated correctly. This example shows that pixel-level depth cues help the model localize and synthesize small objects more reliably.

\noindent\textbf{Ablation on the Reference Image.}
As described in \Cref{sec:3d_transformation_module}, we feed a 3D-transformed reference image to the DiT backbone as a visual preview. We ablate both the reference image and the reference-aware training as shown in \Cref{tab:ablation_reference_image}.

We observe that the reference image is critical. With or without training, using a reference image consistently works better than removing it. The benefit of training is also much larger when the reference image is present. Without the reference image, training only reduces Chamfer from 55.38 to 52.93 (-2.45). With the reference image, training reduces Chamfer from 46.31 to 18.93 (-27.38).

\begin{table}[t]
    \centering
    \caption{Quantitative comparison of the effects of reference image and reference-aware training.}
    \label{tab:ablation_reference_image}
    \begin{tabular}{>{\centering\arraybackslash}w{c}{1cm}>{\centering\arraybackslash}w{c}{1cm}*{4}{>{\centering\arraybackslash}w{c}{1.1cm}}}
    \toprule
        Reference & Training & DIoU$\uparrow$ & AbsRel$\downarrow$ & Chamfer$\downarrow$ & RA-DINO$\uparrow$ \\
    \midrule
        \xmark & \xmark & 45.58 & 68.22 & 55.38 & 18.85 \\
        \xmark & \cmark & 46.16 & 67.99 & 52.93 & 18.87 \\
        \cmark & \xmark & \underline{53.28} & \underline{64.63} & \underline{46.31} & \underline{26.09} \\
        \cmark & \cmark & \textbf{65.33} & \textbf{49.53} & \textbf{18.93} & \textbf{36.91} \\
    \bottomrule
    \end{tabular}
\end{table}

This ablation reflects the challenge of using text-only control for precise 3D manipulation in current open-source frameworks. While upgrading text encoders can improve spatial instruction understanding—as evidenced by the performance gap between Qwen-Image-2.0-Pro~\cite{qwen-team-2026-qwenimage2} and Qwen-Image-Edit~\cite{qwen-image-edit-2511}—fine-grained 3D intent remains difficult to accurately recover from language alone. In our setting, providing a reference image supplies explicit spatial context that text cannot reliably convey, leading to superior results. This text-to-spatial bottleneck is where open models still fall behind leading closed-source systems. Narrowing this gap is crucial for achieving better physical plausibility in complex scene manipulation.

%% file: sections/05-conclusion.tex
\section{Conclusion}
\label{sec:conclusion}

In this paper, we develop a DiT-based image editing framework, \METHODPLACEHOLDER{}, for 3D-aware object manipulation. The framework combines contextual 3D-aware visual guidance with joint 2D-3D supervision to improve physical consistency and spatial accuracy. To support this method and evaluate performance, we present a real-world dataset, \DATASETPLACEHOLDER{}, with detailed object-level annotations, and build a benchmark, \BENCHMARKPLACEHOLDER{}, to evaluate the physical accuracy of object manipulation across multiple dimensions. Experiments show that \METHODPLACEHOLDER{} outperforms existing methods, including strong commercial baselines, on manipulation-related metrics, and ablation studies validate the contribution of each key component. We believe that achieving physically-grounded image editing is a step towards the broader goal of building interactive world models.

%% file: sections/06-appendix.tex
\providecommand{\thirdbest}[1]{\textit{#1}}
\section{Experimental Details}

\subsection{Training Details}

This section adds training details that were omitted from the main paper for space.

\noindent\textbf{Base model and fine-tuning.}
We take Qwen-Image-Edit-2511~\cite{qwen-image-edit-2511} as the backbone; it is a standard choice for image-editing fine-tuning. We add LoRA~\cite{hu2022lora} with rank $128$ mainly to the self-attention blocks of the DiT, and use a smaller rank on the remaining layers (e.g., MLPs).

\noindent\textbf{Prompt template.}
\label{para:prompt_template}
The prompt follows this template:\\
{\small\ttfamily\raggedright
``Assume the image is in a 3D space where the origin is at the top-left-near corner of the image. The X-axis ranges from left (0) to right (1), the Y-axis ranges from top (0) to bottom (1), and the Z-axis (depth) ranges from near (0) to far (1). Move the <OBJECT\_NAME> at <SOURCE\_POSITION> to <TARGET\_POSITION>, and <OBJECT\_EDIT\_INSTRUCTIONS>.\\
<ADDITIONAL\_INSTRUCTIONS>.''\par
}
Our pairs come from real videos, so edits are not limited to the object positions. We use an LLM~\cite{qwen3.5} to compare the source and target images and generate the {\small\ttfamily<OBJECT\_EDIT\_INSTRUCTIONS>} and {\small\ttfamily<ADDITIONAL\_INSTRUCTIONS>} for changes beyond pure relocation.

\noindent\textbf{Depth estimation.}\par
\noindent Depth maps are produced with Depth-Anything-3~\cite{depthanything3}. Single-view depth estimation is not reliable enough for our setting, so at training time we always feed the model an image pair from the same edit. The edited image is paired with the source to obtain $\mathbf{D}_{\text{edit}}$; the target is paired with the source to obtain the target depth $\mathbf{D}_{\text{tgt}}$. In symbols,
\begin{gather}
\mathbf{D}_{\text{src}}, \mathbf{D}_{\text{edit}} = \operatorname{DA3}([\mathbf{I}_{\text{src}}, \mathbf{I}_{\text{edit}}]),\\
\mathbf{D}_{\text{src}}, \mathbf{D}_{\text{tgt}} = \operatorname{DA3}([\mathbf{I}_{\text{src}}, \mathbf{I}_{\text{tgt}}]),
\end{gather}
where $\mathbf{D}_{\text{edit}}$, $\mathbf{D}_{\text{src}}$, and $\mathbf{D}_{\text{tgt}}$ are depth estimates for the edited, source, and target frames, and $\mathbf{I}_{\text{src}}$, $\mathbf{I}_{\text{edit}}$, $\mathbf{I}_{\text{tgt}}$ are the corresponding RGB inputs. Depth supervision uses $\mathbf{D}_{\text{edit}}$ against $\mathbf{D}_{\text{tgt}}$.

Depth is predicted at roughly the source resolution. We use the same paired-input procedure at evaluation time.

\subsection{Experimental Settings for Baselines}

We group baselines by how spatial control is given:

\noindent\textbf{(1) Explicit 2D manipulation.}
These methods expect explicit 2D cues. We supply masks, coordinates, and bounding boxes as required. This set includes \textit{GoodDrag}~\cite{zhang2025gooddrag}, \textit{LightningDrag}~\cite{shi2024lightningdrag}, \textit{OBJect 3DIT}~\cite{michel2023object3ditlanguageguided3daware}, \textit{Move-and-Act}~\cite{jiang2024actenhancedobjectmanipulation}, and \textit{PixelMan}~\cite{jiang2025pixelman}.

\noindent\textbf{(2) Explicit 3D manipulation.}
These methods take 3D-style controls (e.g., 3D coordinates or camera pose). We provide the 3D inputs each implementation needs. This set includes \textit{GeoDiffuser}~\cite{sajnani2024geodiffuser} and \textit{DiffusionHandles}~\cite{pandey2024diffusionhandles}.

\noindent\textbf{(3) Implicit editing.}
These systems do not accept explicit spatial handles. We spell out 3D spatial information in text using the template in \Cref{para:prompt_template}, and we add source images marked with the object boxes and overlays for the target regions to guide the model to understand the editing context and perform the editing. This set includes \textit{ChronoEdit}~\cite{wu2025chronoedit}, \textit{GPT-Image-1.5}~\cite{openai-2025-chatgpt-images}, \textit{Qwen-Image-2.0-Pro}~\cite{qwen-team-2026-qwenimage2}, and \textit{Nano Banana Pro}~\cite{google-deepmind-2025-nanobanana-pro}.

Most explicit-control baselines do not accept an extra free-text prompt beyond spatial inputs. So we omit any additional instructions for all baselines. For a baseline that only supports manipulating one object at a time, we run it iteratively until all objects are handled. Overall, we tailor inputs for each baseline and tune its interface so it receives as complete spatial information as its API allows.

\begin{figure*}[t]
    \centering
    \includegraphics[width=\textwidth]{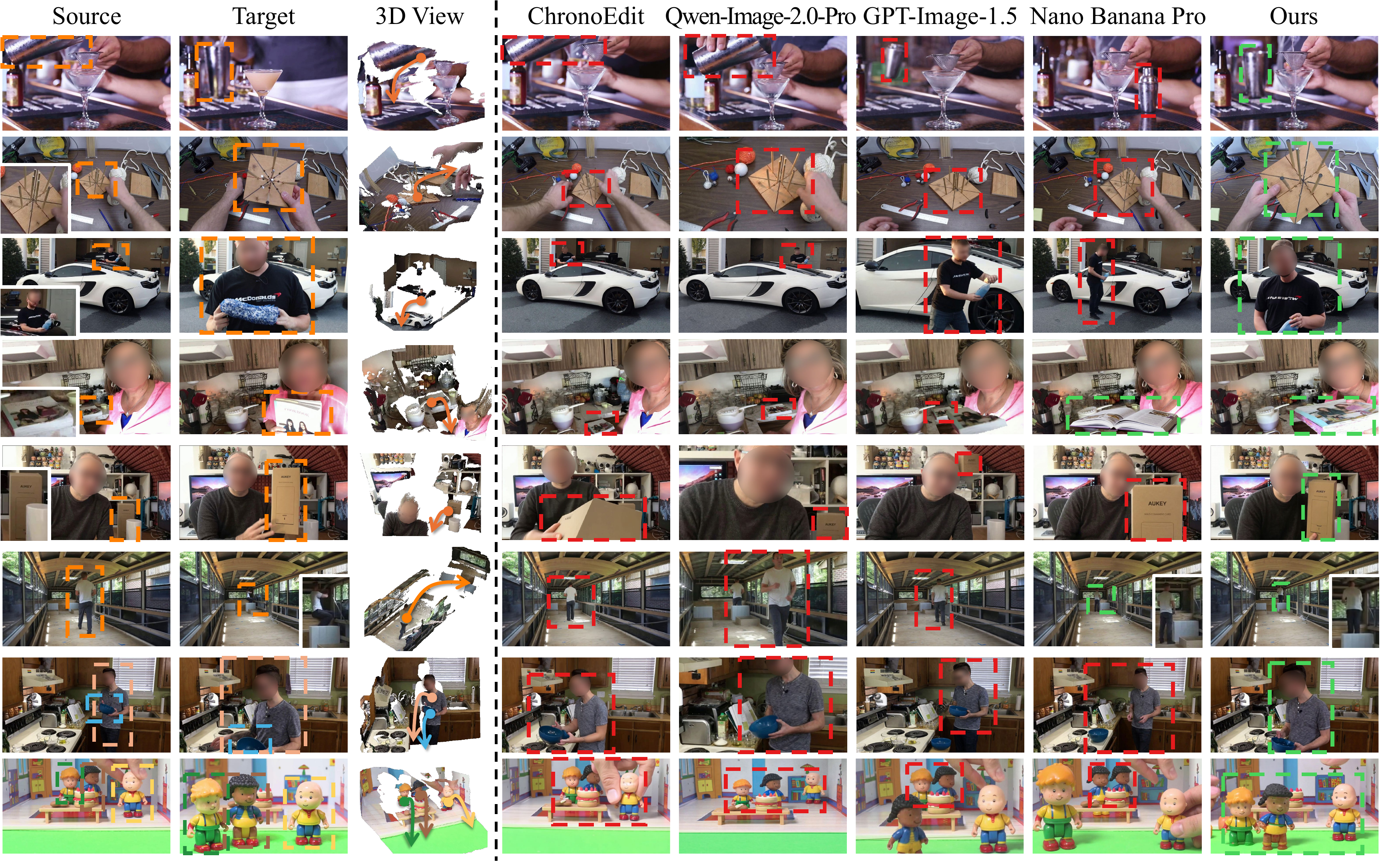}
    \caption{More qualitative results on \BENCHMARKPLACEHOLDER{}. For baseline methods, misplaced objects and objects still left at the source are marked with {\color[HTML]{E51B1D}red} boxes, and correctly manipulated objects are indicated by {\color[HTML]{47D45A}green} boxes.}
    \label{fig:bench_demo_appendix}
    \Description{Eight qualitative comparisons arranged by row, with columns for the source, target, 3D view, four baseline methods, and PhyEdit. The examples move objects including a bottle, a wooden assembly, people, boxes, cookware, and multiple toys. Orange boxes mark source and target regions, red boxes mark incorrect or incomplete edits, and green boxes mark successful placements.}
\end{figure*}

\subsection{Details on Metrics}

\noindent\textbf{Localizing manipulated objects.}
Locating the moved object in generated images is challenging. Feature-based matching~\cite{shi2023dragdiffusion} is too coarse for grounding, and generic detectors~\cite{carion2025sam3segmentconcepts} can be unreliable on edited content. Some baselines may leave a copy at the source, which can confuse the detectors and lead to suboptimal object localization. We use a two-stage procedure: Qwen-3.5-Plus~\cite{qwen3.5} proposes coarse boxes with multimodal reasoning, and SAM3~\cite{carion2025sam3segmentconcepts} refines them to masks.

We pass the ground-truth boxes from the benchmark dataset into the prompt and ask Qwen-3.5-Plus to look for the manipulated object near the target region first, then outside it, and to return a short phrase that uniquely describes that object. SAM3 then refines the coarse box with that phrase to produce the mask.

\noindent\textbf{Penalizing missing objects.}
If the VLM cannot find the object or the mask is empty---sometimes happens when the edit removes the object or collapses to severe artifacts---we still count the sample instead of dropping it. Accuracy-style scores are set to zero for that case. Distance-style scores (Chamfer, centroid, etc.) use a data-driven penalty derived from the global distribution of successful localization:
\begin{equation}
    p = \max(q_{0.99}, 1.2 \cdot q_{0.95}),
\end{equation}
where $q_{\tau}$ denotes the $\tau$-quantile of the empirical metric distribution over successfully localized manipulated objects.

\noindent\textbf{Phys-VLM scoring.}
For each edited image, the VLM receives both the result and a copy annotated with manipulated-object boxes, and assigns a score from 1 to 5. It jointly considers lighting and material consistency, shadows and reflections, contact and edge blending, perspective and scale, depth ordering and occlusion, and global scene coherence. A score of 1 indicates severe artifacts or physical violations, 3 denotes an overall plausible result with noticeable inconsistencies, and 5 is reserved for highly realistic and physically coherent edits. We average the image-level scores for reporting.

\section{Additional Analysis and Validation}

\subsection{Robustness to the 3D Foundation Model}

The 3D transformation module is not tied to a particular geometry backbone. We keep the editor checkpoint and the \BENCHMARKPLACEHOLDER{} evaluation protocol fixed, and replace only the off-the-shelf 3D model used to construct the preview. Neither the 3D backbones nor the editor are further fine-tuned for this comparison. As shown in \Cref{tab:backbone_robustness}, all tested variants retain a positive average geometric improvement over Nano Banana Pro, although the stronger DA3 prior gives the best overall result. This indicates that the method benefits from better geometry but does not rely exclusively on a single estimator.

\begin{table}[t]
    \centering
    \caption{Robustness to different 3D foundation models. Scores are normalized to $[0,100]$. Geo. $\Delta$ is the average signed improvement over Nano Banana Pro across the four geometry metrics after aligning metric directions.}
    \label{tab:backbone_robustness}
    \resizebox{\columnwidth}{!}{%
    \begin{tabular}{lccccc}
        \toprule
        Method & DIoU$\uparrow$ & AbsRel$\downarrow$ & $\delta_{1.25}$$\uparrow$ & Chamfer$\downarrow$ & Geo. $\Delta$$\uparrow$ \\
        \midrule
        Nano Banana Pro & 59.97 & 55.02 & 46.11 & 25.33 & +0.00 \\
        Ours (DA2) & 61.26 & 51.53 & 47.93 & 22.80 & +2.28 \\
        Ours (MoGe) & 60.68 & 49.81 & \textbf{51.14} & 21.82 & +3.62 \\
        Ours (VGGT) & 60.92 & \textbf{48.95} & 48.11 & 21.62 & +3.18 \\
        Ours (Pi3X) & 62.71 & 51.21 & 50.21 & 20.23 & +3.94 \\
        Ours (DA3) & \textbf{65.33} & 49.53 & 51.08 & \textbf{18.93} & \textbf{+5.55} \\
        \bottomrule
    \end{tabular}}
\end{table}

\subsection{Preview Error Audit}

\begin{figure*}[t]
    \centering
    \includegraphics[width=0.96\textwidth]{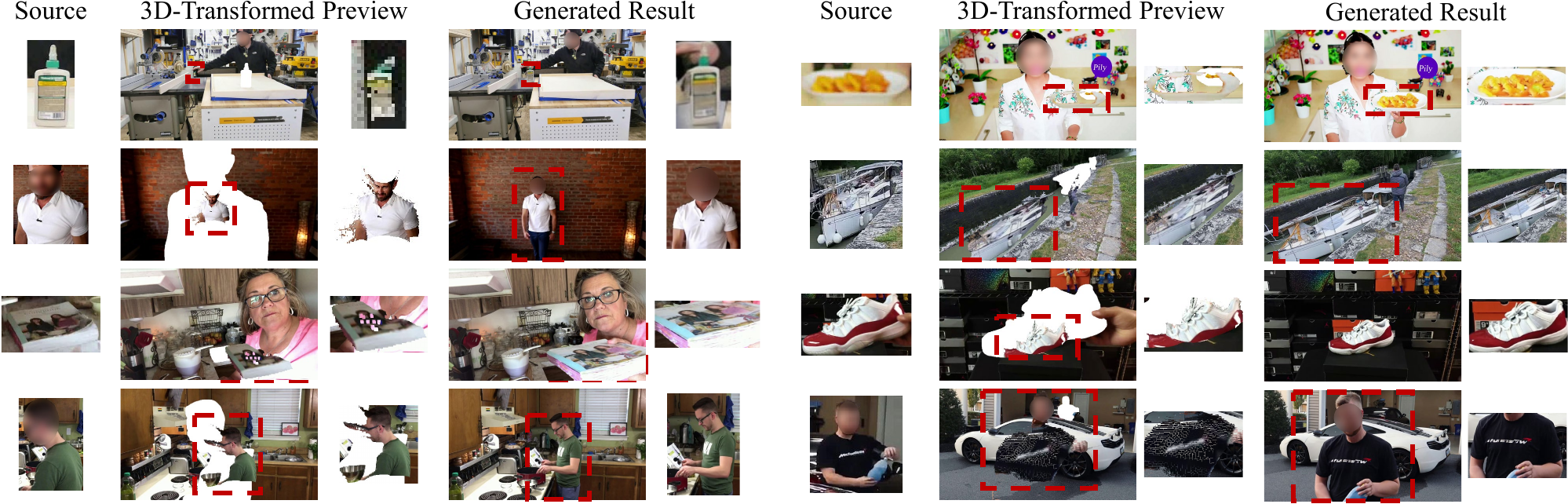}
    \caption{Representative imperfect 3D-transformed previews and their corresponding final outputs. Red dashed boxes mark the manipulated regions. The examples cover boundary and mask errors, distortion under large depth changes, occlusion-order errors, and noisy or fragmented projected surfaces.}
    \label{fig:preview_failure_modes}
    \Description{Pairs of imperfect 3D-transformed previews and final edited images for several object-manipulation cases.}
\end{figure*}

The preview can inherit errors from depth, camera, or mask estimation and from sparse point projection. \Cref{fig:preview_failure_modes} illustrates common artifacts, including incomplete boundaries, missing regions under large depth changes, incorrect occlusion ordering, and fragmented surfaces. The final editor can often use the source image to restore appearance while following the coarse target geometry, but the preview remains an important condition rather than an error-free target. Severe geometry or segmentation failures can therefore propagate to the output, as discussed further in \Cref{sec:appendix_limitations}.

\subsection{Blind Human Evaluation}

\BENCHMARKPLACEHOLDER{} and the training data are disjoint; no evaluation example is used for training.

We conduct a blind pairwise evaluation on a 50-case subset of \BENCHMARKPLACEHOLDER{}. Each case is compared against Nano Banana Pro and Qwen-Image-2.0-Pro, producing 100 anonymous A/B tasks. Ten annotators each evaluate 30 tasks, so every task receives three independent ratings and the study contains 300 task-level ratings. Method identity and left/right ordering are hidden. For each pair, annotators use a five-level choice (A clearly better, A slightly better, tie/uncertain, B slightly better, or B clearly better) on placement and scale, object identity, contact and occlusion, and overall plausibility.

The preference rate in \Cref{tab:human_preference} is computed as $(N_{\text{ours}}+0.5N_{\text{tie}})/N_{\text{all}}$. The ``Overall'' column is a separate overall-plausibility judgment, not an average of the other criteria. PhyEdit is preferred across all four criteria against both baselines, including a $78.0\%$ overall preference against Nano Banana Pro.

\begin{table}[t]
    \centering
    \caption{Blind human preference. Entries are PhyEdit preference rates (\%) over each baseline, with ties assigned half weight.}
    \label{tab:human_preference}
    \setlength{\tabcolsep}{3.2pt}
    \begin{tabular}{lcccc}
        \toprule
        Baseline & Place/Scale & Identity & Contact & Overall \\
        \midrule
        Nano Banana Pro & 81.3 & 69.0 & 74.0 & 78.0 \\
        Qwen-Image-2.0-Pro & 85.3 & 78.7 & 83.0 & 85.3 \\
        \bottomrule
    \end{tabular}
\end{table}

We additionally compare the direction of task-level automatic-score differences with the direction of annotator-averaged judgments. After excluding comparisons tied under either signal, Phys-VLM agrees with contact and occlusion judgments in $33/34$ comparisons ($97.1\%$). RA-DINO agrees with the mean of placement and identity judgments in $83/95$ comparisons ($87.4\%$). The Phys-VLM denominator is small because its discrete score ties on 66 of the 100 tasks. We therefore treat these results as directional sanity checks that complement the human preferences, not as a complete validation of either automatic metric.

\subsection{General Editing During Manipulation}

PhyEdit is specialized for 3D object manipulation rather than unrestricted text editing. We nevertheless test whether the added manipulation conditioning suppresses the backbone's ability to follow local appearance instructions. The test contains 300 object-specific prompts over 200 \BENCHMARKPLACEHOLDER{} items, with two generations per prompt. It covers color or material changes, surface patterns or marks, and local additions. The PhyEdit output must satisfy the appearance edit and 3D relocation in the same generation. For Qwen-Image-Edit, edit success is measured in an editing-only run, while its 3D metrics come from a separate manipulation-only run; this makes the comparison an ability-retention stress test rather than a claim of identical task difficulty.

\begin{figure*}[t]
    \centering
    \includegraphics[width=0.96\textwidth]{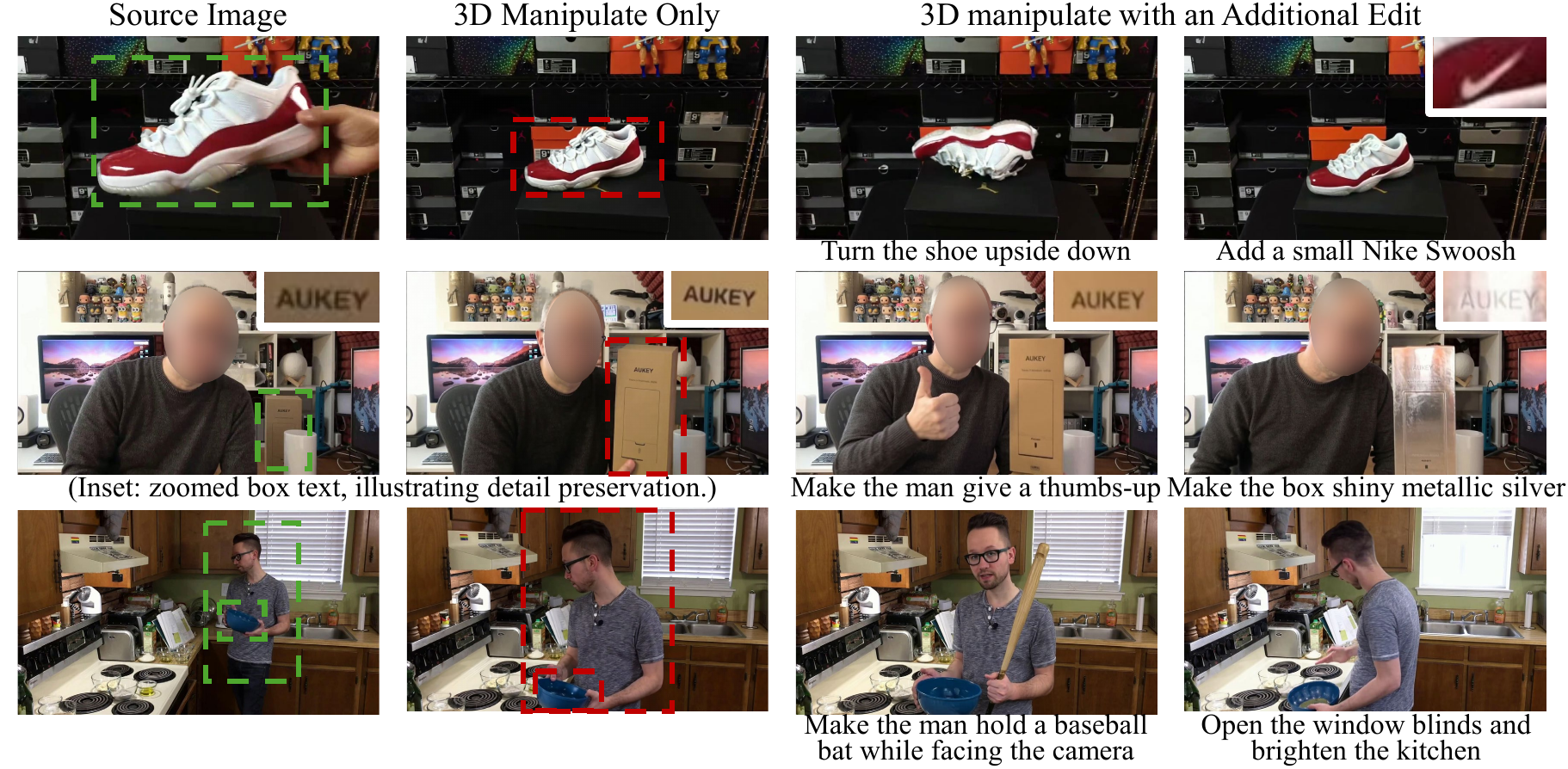}
    \caption{Examples of applying additional appearance or local editing instructions during 3D manipulation. The two added edits in each row are, from top to bottom: flipping the shoe and adding a swoosh; changing the man's gesture and the box material; adding a baseball bat and opening the window blinds. Faces are anonymized except where facial state is necessary to visualize the requested edit.}
    \label{fig:general_editing_demo}
    \Description{Examples in which PhyEdit moves an object in 3D while also changing appearance, adding a local detail, or modifying a nearby scene attribute.}
\end{figure*}

As shown in \Cref{tab:general_edit_stress}, the overall edit-success difference is $1.3$ points, while PhyEdit retains substantially better 3D accuracy in the joint setting. Qualitative examples are shown in \Cref{fig:general_editing_demo}. Fine-grained text, logos, and small patterns remain more difficult and are outside the main claim of this work.

\begin{table*}[t]
    \centering
    \caption{General-edit success (\%) and 3D accuracy. Qwen-Image-Edit uses separate editing-only and manipulation-only runs; PhyEdit performs both in the same generation.}
    \label{tab:general_edit_stress}
    \setlength{\tabcolsep}{4.0pt}
    \begin{tabular}{lccccccc}
        \toprule
        Method & \multicolumn{4}{c}{General-edit success (\%)} & \multicolumn{3}{c}{3D accuracy} \\
        \cmidrule(lr){2-5}\cmidrule(l){6-8}
        & Color$\uparrow$ & Pattern$\uparrow$ & Local$\uparrow$ & All$\uparrow$ & DIoU$\uparrow$ & AbsRel$\downarrow$ & Chamfer$\downarrow$ \\
        \midrule
        Qwen-Image-Edit & \textbf{89.9} & \textbf{88.6} & \textbf{86.8} & \textbf{88.8} & 53.28 & 64.63 & 46.31 \\
        Ours & 88.6 & 86.4 & 86.1 & 87.5 & \textbf{63.68} & \textbf{47.79} & \textbf{20.97} \\
        \bottomrule
    \end{tabular}
\end{table*}

\subsection{Efficiency and Adaptation Cost}

We profile 200 \BENCHMARKPLACEHOLDER{} inputs on one GPU at $768\times432$, with 28 denoising steps and batch size 1. The first five inputs are used for warm-up, and \Cref{tab:efficiency} reports mean compute time over the remaining 195 images; file I/O is excluded. DA3 depth and camera estimation plus preview projection require $0.152$ seconds per image. The larger end-to-end difference comes primarily from processing the additional visual condition in the editor. Peak VRAM increases from $57.7$ to $59.0$ GB.

\begin{table}[t]
    \centering
    \caption{Single-GPU inference profile. Runtime is the mean compute time per image; memory is peak VRAM.}
    \label{tab:efficiency}
    \setlength{\tabcolsep}{4.5pt}
    \begin{tabular}{lcc}
        \toprule
        Component or setting & Runtime (s) & Peak VRAM (GB) \\
        \midrule
        DA3 depth/camera & 0.124 & 9.26 \\
        Preview projection & 0.028 & -- \\
        3D preprocessing total & 0.152 & -- \\
        Base Qwen-Image-Edit & 7.67 & 57.7 \\
        PhyEdit end-to-end & 11.78 & 59.0 \\
        \bottomrule
    \end{tabular}
\end{table}

The base model weights remain frozen during adaptation, and the LoRA modules contain $613.5$M trainable parameters, or $2.92\%$ of the transformer. In matched micro-batch measurements, adding decoded-image depth supervision increases training time from $3.23$ to $4.03$ seconds per micro-batch.

\subsection{More Qualitative Results on \BENCHMARKPLACEHOLDER{}}

Additional qualitative comparisons are shown in \Cref{fig:bench_demo_appendix}.

\noindent\textbf{Manipulation accuracy and consistency.}
In the first two rows most baselines miss the correct placement. \textit{GPT-Image-1.5} (row~1) and \textit{Qwen-Image-2.0-Pro} (row~2) sometimes move the object but mainly rescale it. The result of the manipulation looks like a 2D scale change rather than a true depth shift.

\noindent\textbf{Object identity and fine details.}
In rows~3--6 depth placement is often wrong. In rows~4--5 \textit{Nano Banana Pro} is closer, yet lighting or scale can look slightly unrealistic.

\noindent\textbf{Multi-object editing.}
In the last two rows many baselines fail to handle part of the manipulation requests. Even \textit{Nano Banana Pro} finishes only a subset of objects, whereas our method moves each target as requested consistently.

\begin{table*}[t]
    \caption{Quantitative comparison on \BENCHMARKPLACEHOLDER{} on single-object items, sorted by Chamfer distance in descending order.}
    \centering
    \begin{tabular}{c|*{9}{>{\centering\arraybackslash}w{c}{1.2cm}}}
    \toprule
     \textbf{Method} & DIoU$\uparrow$ & Mask IoU$\uparrow$ & AbsRel$\downarrow$ & $\delta_{1.25}$$\uparrow$ & Chamfer$\downarrow$ & Centroid$\downarrow$ & RA-DINO$\uparrow$ & DeQA$\uparrow$ & Phys-VLM$\uparrow$ \\
    \midrule
    GeoDiffuser & 51.36 & 18.70 & 85.67 & 24.72 & 93.34 & 73.95 & 15.40 & 60.23 & 66.08 \\
    Object 3DIT & 45.34 & 10.03 & 87.47 & 26.78 & 91.11 & 77.96 & 14.64 & 48.70 & 69.70 \\
    Move-and-Act & 50.54 & 12.65 & 83.88 & 26.76 & 73.83 & 68.99 & 15.32 & 64.20 & 66.70 \\
    LightningDrag & 51.99 & 17.94 & 74.68 & 27.84 & 66.65 & 64.31 & 21.93 & 72.80 & \thirdbest{92.00} \\
    Qwen-Image-Edit & 51.41 & 12.70 & 86.77 & \thirdbest{36.54} & 66.43 & 64.38 & 25.56 & \underline{74.76} & \underline{92.80} \\
    GoodDrag & 51.05 & 13.65 & 88.85 & 31.55 & 65.55 & 67.10 & 22.69 & 74.20 & 91.50 \\
    DiffusionHandles & \thirdbest{58.73} & \thirdbest{20.21} & \underline{67.91} & 27.72 & 65.54 & 58.20 & 20.80 & 64.25 & 52.70 \\
    PixelMan & 50.10 & 11.81 & 84.13 & 34.65 & 62.98 & 66.13 & 22.53 & 72.61 & 83.30 \\
    ChronoEdit & 48.79 & 9.28 & 85.58 & 32.55 & 58.09 & 62.37 & 23.82 & 73.87 & 91.85 \\
    GPT-Image-1.5$^\dagger$ & 51.57 & 11.55 & 91.56 & 33.31 & 50.82 & 60.91 & 23.62 & 74.02 & 87.24 \\
    Qwen-Image-2.0-Pro$^\dagger$ & 56.34 & 15.63 & 69.77 & 36.20 & \thirdbest{39.11} & \thirdbest{49.68} & \thirdbest{30.98} & 67.11 & 81.21 \\
    Nano Banana Pro$^\dagger$ & \underline{63.35} & \underline{22.40} & \thirdbest{69.16} & \underline{44.20} & \underline{26.45} & \underline{34.91} & \underline{38.49} & \textbf{76.37} & 90.20 \\
    \midrule
    Ours & \textbf{66.29} & \textbf{29.10} & \textbf{66.66} & \textbf{46.11} & \textbf{21.26} & \textbf{34.45} & \textbf{39.14} & \thirdbest{74.37} & \textbf{93.23} \\
    \bottomrule
    \end{tabular}
    
    \label{tab:baseline_comparison_single_object}
\end{table*}

\begin{table*}[t]
    \caption{Quantitative comparison on \BENCHMARKPLACEHOLDER{} on multi-object items, sorted by Chamfer distance in descending order.}
    \centering
    \begin{tabular}{c|*{9}{>{\centering\arraybackslash}w{c}{1.2cm}}}
    \toprule
     \textbf{Method} & DIoU$\uparrow$ & Mask IoU$\uparrow$ & AbsRel$\downarrow$ & $\delta_{1.25}$$\uparrow$ & Chamfer$\downarrow$ & Centroid$\downarrow$ & RA-DINO$\uparrow$ & DeQA$\uparrow$ & Phys-VLM$\uparrow$ \\
    \midrule
    DiffusionHandles & 53.72 & 17.55 & 46.81 & 35.74 & 79.35 & 59.70 & 13.35 & 59.42 & 41.10 \\
    GeoDiffuser & 52.07 & \underline{21.49} & 49.79 & 38.01 & 61.58 & 55.14 & 14.17 & 53.24 & 43.27 \\
    Move-and-Act & 43.25 & 9.29 & 53.05 & 31.85 & 56.91 & 56.97 & 10.69 & 57.39 & 55.00 \\
    Object 3DIT & 47.92 & 10.33 & 46.39 & 38.72 & 40.29 & 51.06 & 13.99 & 47.31 & 63.60 \\
    GoodDrag & 51.08 & 12.92 & 50.16 & 36.85 & 37.65 & 49.17 & 18.98 & 73.96 & 78.59 \\
    PixelMan & 48.87 & 11.17 & 48.79 & 35.55 & 37.62 & 47.63 & 19.81 & 72.96 & 63.90 \\
    ChronoEdit & 49.05 & 9.24 & 46.02 & 40.16 & 32.62 & 46.39 & 19.65 & \thirdbest{76.71} & \underline{92.45} \\
    LightningDrag & 55.63 & \thirdbest{19.86} & \underline{39.45} & 44.13 & 29.81 & 43.44 & 21.91 & 74.41 & 85.20 \\
    Qwen-Image-Edit & 55.14 & 14.91 & 42.49 & 44.68 & 28.97 & 41.02 & 26.62 & 76.58 & 88.30 \\
    GPT-Image-1.5$^\dagger$ & 53.09 & 11.14 & 50.55 & 39.65 & 28.87 & 44.72 & 23.97 & \underline{77.61} & 90.10 \\
    Nano Banana Pro$^\dagger$ & \thirdbest{56.55} & 15.43 & \thirdbest{40.74} & \underline{48.04} & \thirdbest{23.75} & \underline{36.13} & \thirdbest{31.00} & \textbf{78.61} & \thirdbest{91.92} \\
    Qwen-Image-2.0-Pro$^\dagger$ & \underline{56.62} & 15.57 & 42.22 & \thirdbest{46.54} & \underline{21.68} & \thirdbest{37.62} & \underline{31.11} & 69.39 & 84.60 \\
    \midrule
    Ours & \textbf{64.38} & \textbf{25.30} & \textbf{32.40} & \textbf{56.06} & \textbf{17.17} & \textbf{29.80} & \textbf{34.68} & \thirdbest{76.58} & \textbf{94.21} \\
    \bottomrule
    \end{tabular}
    \label{tab:baseline_comparison_multi_object}
\end{table*}

\subsection{Results on \BENCHMARKPLACEHOLDER{} Split by Object Number}

Multi-object scenes in \BENCHMARKPLACEHOLDER{} contain two to four edited objects. Compared to the single-object split, depth changes are not large for every object in a scene; however, at least one object has a clear depth shift, and the average depth change remains quite substantial.

Quantitative results for each split are in \Cref{tab:baseline_comparison_single_object,tab:baseline_comparison_multi_object}. The two splits are not directly comparable, but the tables suggest the gap versus strong baselines can shift with the number of objects.

On the single-object split, our method leads on the manipulation metrics in \Cref{tab:baseline_comparison_single_object}; exact values are in the table. Relative to Nano Banana Pro, Chamfer distance improves by 5.19 points and Absolute Relative Error decreases by 2.50 points, with consistent gains also on DIoU, Mask IoU and Phys-VLM.

On the multi-object split (\Cref{tab:baseline_comparison_multi_object}), the ordering changes, and the same comparison against Nano Banana Pro shows slightly larger differences: Chamfer from 5.19 to 6.58, and $\delta_{1.25}$ from 1.91 to 8.02. This is consistent with the fact that multi-object scene edits require more precise depth and placement consistency. In this setting, our method maintains better physical accuracy and a clearer advantage.

Leading closed-source models, especially Nano Banana Pro, can generate high-quality images. But as shown in the qualitative examples above, its results can still have unrealistic lighting or object scale after manipulation, which hurts physical realism. Phys-VLM shows a similar trend across the two splits.

Overall, this split-by-object-number analysis shows that our method keeps an advantage on geometry-sensitive metrics in both settings, and that this advantage is more pronounced in the multi-object case.

\section{Limitations}
\label{sec:appendix_limitations}

\noindent\textbf{Insufficient prompt control.}\par
\noindent As discussed in \Cref{para:prompt_template}, changes in real-video image pairs are not always fully aligned with the additional prompt and the text may not cover all visible edits. In addition, our current training does not include dedicated optimization for the additional free-form text branch. As a result, prompt control from free-text additional instructions is still not that strong, especially for fine-grained constraints about where to edit and where to preserve content.

\textbf{Failure to handle extreme cases.}
Our method can still fail in difficult scenes. Artifacts are more likely when the request is highly complex, the scene is heavily cluttered, or the target motion is geometrically extreme (e.g., moving an object too close to the camera, where parts may fall behind the lens, or where projected geometry becomes too dense). Likewise, when an object is moved from far distance to near distance, appearance details may not be recovered reliably.

\textbf{Dependence on upstream geometry.}
Depth, camera, and mask errors directly affect the transformed preview. The backbone comparison and qualitative audit show some tolerance to imperfect previews, but severe errors, transparent or reflective objects, and texture-poor regions remain failure cases. The explicit module models rigid relocation, projection, scale, and depth ordering rather than forces, collisions, or dynamics; depth supervision likewise does not ensure complete geometric or physical correctness.